\documentclass[sigconf]{acmart} 

\AtBeginDocument{%
  }

\setcopyright{acmlicensed}
\copyrightyear{2024}
\acmYear{2024}
\acmDOI{XXXXXXX.XXXXXXX}

\acmISBN{978-1-4503-XXXX-X/18/06}
\acmSubmissionID{69}

\usepackage{graphicx}
\usepackage{amsmath}
\usepackage{amsthm}
\usepackage{booktabs}
\usepackage{algorithm}
\usepackage{algorithmic}
\usepackage{graphicx,subfigure}
\usepackage{longtable}
\usepackage{rotating}
\usepackage{multirow}
\usepackage{makecell}
\usepackage{bm}
\usepackage{float}
\usepackage{flushend}
\usepackage{times}
\usepackage{xcolor}
\usepackage{color}
\usepackage{soul}
\usepackage{tabularx}
\usepackage{amsfonts}
\usepackage{epstopdf}
\usepackage{latexsym}
\usepackage{setspace}
\newtheorem{theorem}{Theorem}
\newtheorem{remark}{Remark}

\newtheorem{definition}{Definition}
\def\Sp{{\scriptsize{\textcircled{{\emph{\tiny{\textbf{Sp}}}}}}}}

\begin{document}

\title{One-Step Multi-View Clustering Based on Transition Probability}
\author{Wenhui Zhao, Quanxue Gao, Guangfei Li, Cheng Deng, Ming Yang}


\renewcommand{\shortauthors}{Trovato et al.}

\begin{abstract}
The large-scale multi-view clustering algorithms, based on the anchor graph, have shown promising performance and efficiency and have been extensively explored in recent years. Despite their successes, current methods lack interpretability in the clustering process and do not sufficiently consider the complementary information across different views. To address these shortcomings, we introduce the One-Step Multi-View Clustering Based on Transition Probability (OSMVC-TP). This method adopts a probabilistic approach, which leverages the anchor graph, representing the transition probabilities from samples to anchor points. Our method directly learns the transition probabilities from anchor points to categories, and calculates the transition probabilities from samples to categories, thus obtaining soft label matrices for samples and anchor points, enhancing the interpretability of clustering. Furthermore, to maintain consistency in labels across different views, we apply a {\textit{Schatten p}}-norm constraint on the tensor composed of the soft labels. This approach effectively harnesses the complementary information among the views. Extensive experiments have confirmed the effectiveness and robustness of OSMVC-TP.

\end{abstract}





\maketitle

\section{Introduction}
With the ongoing progress in data generation and feature extraction methodologies, the application of multi-view data has seen a significant rise. Multi-view clustering, a paradigm of unsupervised learning delineated in \cite{bickel2004multi}, is leveraged for grouping data into distinct clusters. This technique is particularly useful in scenarios where explicit labels are unavailable. Predominantly, it has been employed in diverse fields such as image segmentation \cite{sharma2016review}, object recognition \cite{jain1999data}, among others.

In recent years, there has been a substantial development in the domain of multi-view clustering methods. This includes the emergence of subspace-based clustering algorithms \cite{luo2018consistent,zhang2018generalized,zhang2017latent}, non-negative matrix factorization-based clustering algorithms \cite{liu2013multi,yu2021novel,li2016graph}, and graph-based clustering algorithms \cite{wang2019gmc,ZhanNCNZY19,zhan2017graph}. Notably, graph-based clustering algorithms have garnered considerable attention. This is attributed to their efficacy in representing inter-data relationships and unravelling complex hidden structures within the data.

Graph-based methods are pivotal in multi-view clustering, leveraging a discriminative and view-consistent graph while utilizing graph partitioning for clustering. MLAN \cite{NieCLL18} adopts a Laplacian rank constraint to craft a view-consistent graph with distinct, clear connected components, facilitating direct label extraction. SwMC \cite{NieLL17} introduces a novel, non-parametric weight learning strategy, adaptively tuning the weights of different view graphs in consideration of their varied contributions to clustering, markedly enhancing performance. Furthermore, ETLMSC \cite{wu2019essential} integrates tensor nuclear norm constraints, optimizing the synergy of complementary inter-view information for superior view-consistent graph quality.

Despite their successes, these methods generally construct \( N \times N \) graphs, a strategy that becomes inefficient with large-scale multi-view datasets. Addressing time and space efficiency, anchor graph-based methods, such as those proposed in \cite{liu2010large}, have gained traction. These techniques involve selecting \( M \) representative data points as anchors from \( N \) samples, balancing global data representation and computational simplicity, thereby boosting clustering efficiency. Typical of such methods is the use of an anchor graph to formulate a bipartite graph, from which clustering labels are derived based on view-consistent connectivity. MVSC \cite{li2015large}, for instance, excels in managing large-scale multi-view data, but demands high-quality pre-defined bipartite graphs. SFMC \cite{9146384} merges bipartite graph learning with Laplacian rank constraints, aiming for clear component segregation in the learned bipartite graphs. Additionally, TBGL \cite{XiaGWGDT23} employs Schatten p-norm regularization on the tensor of bipartite graphs, effectively harnessing complementary inter-view data and ensuring view consistency and label uniformity. PAGG \cite{0002RYLY23} enhances this approach by introducing predefined anchor point labels, imposing constraints on anchor points to foster discriminative cluster structures and equitable view allocation, thereby yielding superior bipartite graphs for clustering. However, these methods often lack interpretability and necessitate precise model parameter settings, complicating the attainment of \( K \) distinct connected components.

Based on the identified limitations of current approaches, we propose a novel technique named \textit{One-Step Multi-View Clustering Based on Transition Probability}. This method aims to enhance the explanatory power of clustering analysis. It leverages the concept that the anchor graph is analogous to the transition probability matrix from samples to anchor points, as described in \cite{liu2010large}. Our key idea is to determine the transition probability matrix from anchor points to categories. By doing so, we can directly derive the transition probability from samples to categories within a probabilistic framework. This approach enables us to learn concurrently the transition probability matrix from anchor points to categories and a soft label matrix for the samples. The reliability of our results is further bolstered by measuring the discrepancy between the calculated transition probability from samples to categories and the inferred soft label matrix using the Frobenius norm. Moreover, the transition probability matrix can serve as a soft label matrix for the anchor points. Despite potential significant differences in data distributions across various views, their inherent structures are expected to remain consistent. Consequently, we posit that the labels for both samples and anchor points should be consistent across different views. To achieve this, we introduce a \textbf{\textit{Schatten p}}-norm constraint \cite{gao2020tensor} on both the transition probability matrix and the soft label matrix. This constraint facilitates the extraction of complementary information between views, thereby yielding more accurate clustering labels.

The key contributions of this study are summarized as follows:
\begin{itemize}
	\item We adopt a probabilistic perspective to analyze the relationship between the anchor graph and the soft label matrix, assigning meaningful probability associations to improve the interpretability of the clustering model.
	\item The application of the \textbf{\textit{Schatten p}}-norm constraint allows us to effectively harness complementary information across different views. This ensures consistency in the labels of samples and anchor points across these views, thus enhancing the overall clustering performance.
	\item Extensive experiments conducted on four small-scale datasets and two large-scale datasets validate the effectiveness of our proposed method.
\end{itemize}

{\emph{Notations:}} Throughout this article, we use bold upper case letters for matrices, \emph{e.g.}, ${\mathbf{D}}$; bold lower case letters for vectors, \emph{e.g.}, ${\bf{d}}$, lower case letters denotes the entry of ${\bm{\mathcal {D}}}$, \emph{e.g.}, ${D_{ijk}}$, the entry of $\textbf{D}$ is $D_{ij}$. Besides, bold calligraphy letters $\bm{\mathcal{D}} \in{\mathbb{R}} {^{{n_1} \times {n_2} \times {n_3}}}$ for 3-order tensor; the $i$-th frontal slice of ${\bm{\mathcal {D}}}$ is ${\bm{\mathcal {D}}}^{(i)}$. $\overline {{\bm{\mathcal {D}}}}$ is the discrete Fourier transform (DFT) of ${\bm{\mathcal {D}}}$ along the third dimension, $\overline {{\bm{\mathcal {D}}}} = \mathrm{fft}({{\bm{\mathcal D}}},[\ ],3)$. Thus, $\bm{{\mathcal D}} = \mathrm{ifft}({\overline {\bm{\mathcal D}}},[\ ],3)$. $\textbf{I}$ is an identity matrix.

\section{Probability Transition Process}
\begin{figure}[!t]
	\centering
	\includegraphics[width=0.7\linewidth]{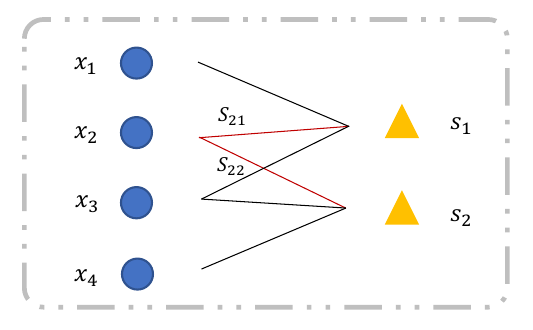}
	\caption{An anchor graph representation of samples and anchors.}
	\label{anchorfigure}
\end{figure}
The properties of an anchor graph $\mathbf{S}$ are characterized as follows: 
\begin{enumerate}
    \item Each element in $\mathbf{S}$ is non-negative.
    \item The sum of elements in each row of $\mathbf{S}$ equals 1.
\end{enumerate}
Hence, as illustrated in Fig.~\ref{anchorfigure}, $\mathbf{S}$ delineates the interconnections between $n$ samples and $m$ anchors, where $\mathbf{S}_{ij}$ represents the affinity of the $i$-th sample to the $j$-th anchor, satisfying $\mathbf{S}_{21} + \mathbf{S}_{22} = 1$. 

We define stationary Markov random walks on $\mathbf{S}$, following Dynkin's framework \cite{dynkin1965markov}. The one-step transition probability from the $i$-th sample to the $j$-th anchor is given by:
\begin{equation}
    p^{(1)} (s_j|x_i) = \frac{S_{ij}}{\sum_{j'}^{m} S_{ij'}} = S_{ij}
\end{equation}

Additionally, the transition from anchor points to categories is modeled as a Markov process, where the transition probability is contingent solely on the current state. Given the one-step transition probability matrix from anchor points to categories as $\mathbf{H}$, we have:
\begin{equation}
    p^{(1)} (c_k|s_j) = H_{kj}
\end{equation}

Since the two Markov processes are independent, the overall transition probability from samples to categories is computed as:
\begin{equation}
    p(c_k|x_i) = \sum_{j=1}^{m} p^{(1)} (c_k|s_j) p^{(1)} (s_j|x_i)
\end{equation}

By calculating these probabilities, we determine the likelihood of each sample belonging to various categories. The category with the highest probability is deemed the most probable label. This label assignment completes the label transition process. Moreover, by transforming these computations into matrix operations, the product $\mathbf{SH}$ yields a matrix representing the transition probabilities from each sample to each category. This matrix, interpreted as the soft label matrix for the samples, facilitates direct acquisition of sample labels.

\section{Methodology}
\subsection{Motivation and Objective Function}
\begin{figure*}
	\centering
	\includegraphics[width=0.8\linewidth]{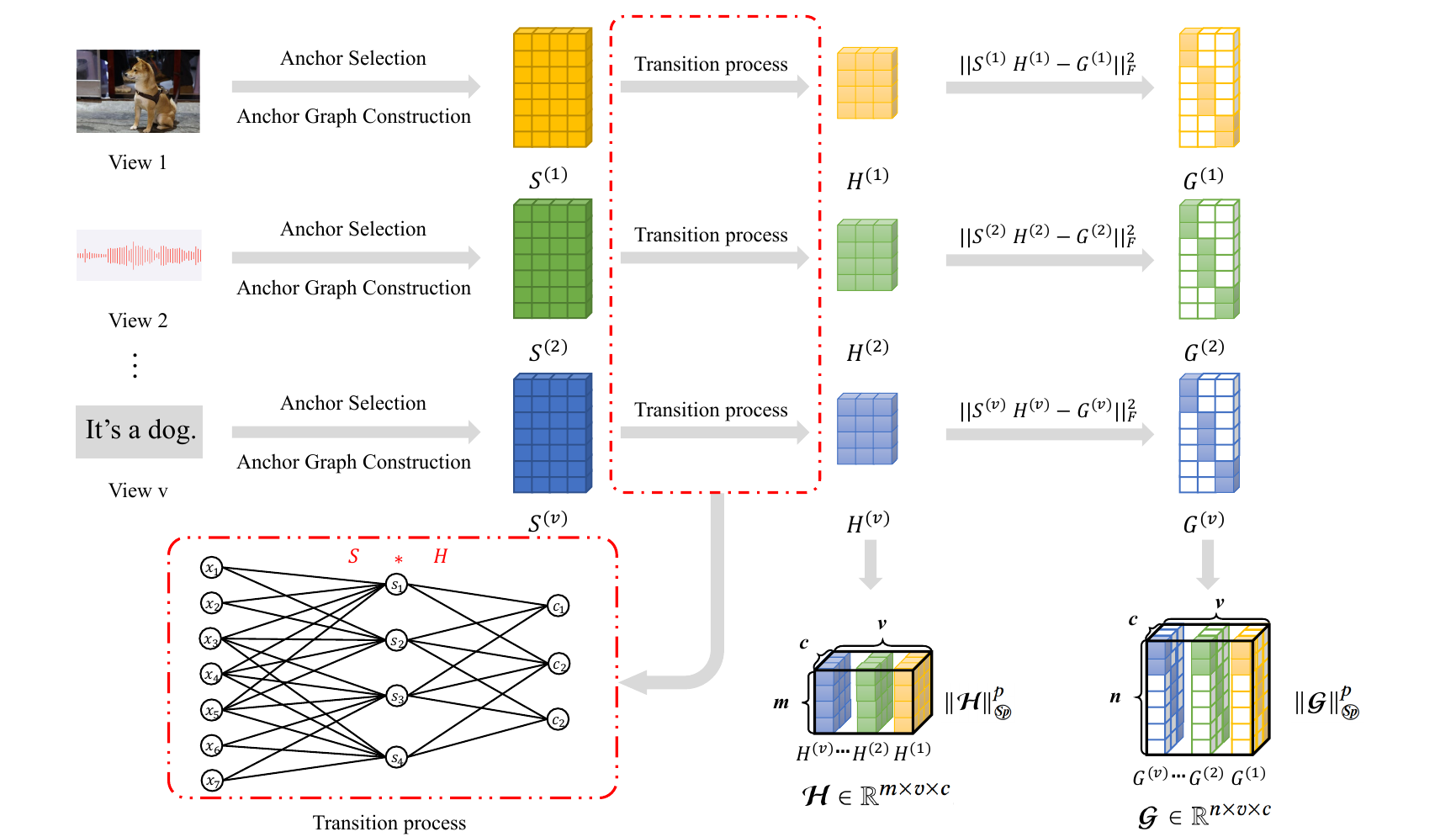}
	\caption{The anchor graph $\mathbf{S}^{(v)}$ can represent the transition probabilities from samples to anchor points. If the transition probabilities from anchor points to categories $\mathbf{H}^{(v)}$ are known, it is possible to directly compute the transition probabilities from samples to categories. The \textbf{\textit{Schatten p}}-norm is applied to complementary information.}
	\label{model}
\end{figure*}
Most existing large-scale multi-view clustering algorithms reduce algorithmic complexity by relying on anchor graphs. By selecting $m$ representative anchor points that cover the entire distribution within the dataset, the internal structure of the data can be represented. An anchor graph can be constructed based on the relationship between $n$ sample points and $m$ anchor points, and when a sample and an anchor point belong to the same category, their correlation is stronger. Given that an anchor graph possesses non-negative properties and its row sums equal to 1, we can consider the anchor graph as a probability transition matrix from the samples to the anchor points. If we know the transition probability matrix from the anchor points to the categories, we can then calculate the transition probability matrix from the samples to the categories in one step. The loss function is:
\begin{equation}
	\begin{aligned}	
		&\mathop{\min_{\mathbf{G}^{(v)},\mathbf{H}^{(v)}}}\sum_{v=1}^{V}
		\frac{1}{\alpha^{(v)}} \|
		\mathbf{S}^{(v)}\mathbf{H}^{(v)} -
		\mathbf{G}^{(v)}\|_{F}^{2} + \lambda\mathcal{R}(\mathbf{G}^{(v)},\mathbf{H}^{(v)})
		\\&s.t.\;\mathbf{G}^{(v)}\geq 0,\;
		\mathbf{G}^{(v)}\cdot 1 =1,\;
		\mathbf{H}^{(v)}\geq 0,\;
		\mathbf{H}^{(v)}\cdot \textbf{1} =\textbf{1}\;
		\\&\qquad\alpha^{(v)} \geq 0,\; \sum_{v=1}^{V}\alpha^{(v)} = 1
	\end{aligned}	
\end{equation}

In this formulation, $\mathbf{S}^{(v)} \in \mathbb{R}^{n \times m}$ represents the predefined anchor graph of the $v$-th view, where $m$ denotes the number of anchors. It satisfies $\mathbf{S}^{(v)} \geq 0$ and $\mathbf{S}^{(v)} \mathbf{1} = \mathbf{1}$, implying that it can be viewed as the probability transition matrix from the samples to the anchor points. $\mathbf{H}^{(v)} \in \mathbb{R}^{m \times c}$ is the probability transition matrix from the anchor points to the categories, and $\mathbf{G}^{(v)} \in \mathbb{R}^{n \times c}$ is the probability transition matrix from the samples to the categories. Hence, $\mathbf{G}^{(v)}$ indicates the categories of the samples, and $\mathbf{H}^{(v)}$ provides information about the categories of the anchor points. $\mathcal{R}(\mathbf{G}^{(v)}, \mathbf{H}^{(v)})$ is a regularization term.

Since we use $\mathbf{G}^{(v)}$ to obtain sample labels, we relax the constraints on $\mathbf{G}^{(v)}$ to be non-negative and orthogonal to facilitate a more intuitive understanding of the categories. In this scenario, each row of $\mathbf{G}^{(v)}$ contains only one non-zero value, and the position of this non-zero value corresponds to the label of the sample. The objective function is:
\begin{equation}
	\begin{aligned}	
		&\mathop{\min_{\mathbf{G}^{(v)},\mathbf{H}^{(v)}}}\sum_{v=1}^{V}
		\frac{1}{\alpha^{(v)}} \|
		\mathbf{S}^{(v)}\mathbf{H}^{(v)} -
		\mathbf{G}^{(v)}\|_{F}^{2} + \lambda\mathcal{R}(\mathbf{G}^{(v)},\mathbf{H}^{(v)})
		\\&s.t.\;\mathbf{G}^{(v)^{\mathrm{T}}}\mathbf{G}^{(v)}=\mathbf{I},\;
		\mathbf{G}^{(v)}\geq 0,\;
		\mathbf{H}^{(v)}\geq 0,\;
		\mathbf{H}^{(v)}\cdot \textbf{1} =\textbf{1}\;
		\\&\qquad\alpha^{(v)} \geq 0,\; \sum_{v=1}^{V}\alpha^{(v)} = 1
	\end{aligned}	
\end{equation}

\begin{figure}
	\centering
	\includegraphics[width=1.0\linewidth]{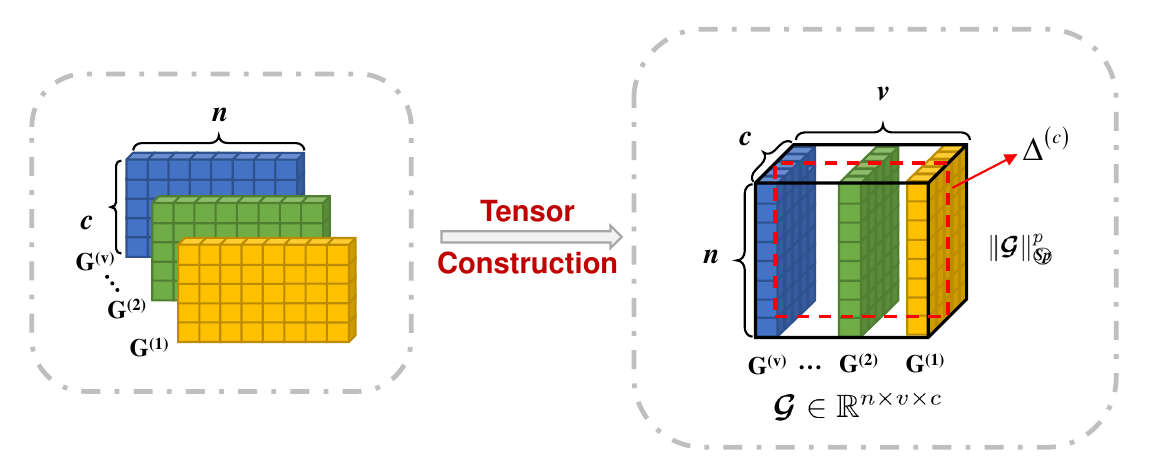}
	\caption{Construction of $\bm{\mathcal{G}} \in \mathbb{R}^{n\times v \times c}$. $\Delta^{(c)}$ is the $c$-th frontal slice of $\bm{\mathcal{G}}$.}
	\label{tensor}
\end{figure}

Moreover, despite the differences in data distribution across different views, 
the fundamental geometric structure remains unchanged. This implies that the 
labels for sample points and anchor points in each view should be consistent. 
In other words, the transition probabilities from each view to their respective 
categories should be uniform. To extract complementary information across views, 
we impose a Schatten p-norm constraint on both matrices $\mathbf{G}^{(v)}$ and 
$\mathbf{H}^{(v)}$. This constraint ensures that the labels for samples and anchors 
within each view are kept consistent. As a result, we formulate the comprehensive 
objective function as follows and the model is shown in Fig. \ref{model}:

\begin{equation}
	\begin{aligned}	
		&\mathop{\min}\sum_{v=1}^{V}
		\frac{1}{\alpha^{(v)}} \|
		\mathbf{S}^{(v)}\mathbf{H}^{(v)} -
		\mathbf{G}^{(v)}\|_{F}^{2} +
		\lambda_1\|\bm{\mathcal{G}}\|_{\Sp}^{p} +
		\lambda_2\|\bm{\mathcal{H}}\|_{\Sp}^{p}
		\\&s.t.\;\mathbf{G}^{(v)^{\mathrm{T}}}\mathbf{G}^{(v)}=\mathbf{I},\;
		\mathbf{G}^{(v)}\geq 0,\;
		\mathbf{H}^{(v)}\geq 0,\;
		\mathbf{H}^{(v)}\cdot \textbf{1} =\textbf{1}\;
		\\&\qquad\alpha^{(v)} \geq 0,\; \sum_{v=1}^{V}\alpha^{(v)} = 1
	\end{aligned}	
\end{equation}
where $\mathbf{G}^{(v)}$ and $\mathbf{H}^{(v)}$ represent the $i$-th lateral slice of tensors $\bm{\mathcal{G}} \in \mathbb{R}^{n\times v \times c}$ and $\bm{\mathcal{H}} \in \mathbb{R}^{m\times v \times c}$, respectively. For example, $\bm{\mathcal{G}}$ is illustrated in Fig. \ref{tensor}. The definition of $\| \bullet \|_{\Sp}$, as described in \cite{gao2020tensor}, is:

\begin{definition}\label{Sp-norm}
	\cite{gao2020enhanced} Let ${\bm{\mathcal D}}\in{\mathbb{R}}^{n_1 \times n_2 \times n_3}$ and $h = \min(n_1,n_2)$. The tensor Schatten $p$-norm, denoted as $\|{\bm{\mathcal D}}\|_{\Sp}^{p}$, is defined as:
	\begin{equation}
		\begin{aligned}
			{\left\| {\bm{\mathcal D}} \right\|_{{\Sp}}} \textrm{=} {\left( {\sum\limits_{i = 1}^{{n_3}} {\left\| {{{\overline {\bm{\mathcal D}} }^{(i)}}} \right\|}_{{\Sp}}^p} \right)^{\frac{1}{p}}} \textrm{=} {\left( {\sum\limits_{i = 1}^{{n_3}} {\sum\limits_{j = 1}^h {{\sigma _j}{{\left( {{{\overline {\bm{\mathcal D}} }^{(i)}}} \right)}^p}} } } \right)^{\frac{1}{p}}}
		\end{aligned}\label{4}
	\end{equation}
	with $0<p \leq 1$. Here, ${\sigma _j}(\overline{\bm{\mathcal D}}^{(i)})$ stands for the j-th singular value of $\overline{\bm{\mathcal D}}^{(i)}$. By selecting an appropriate value of p, we can more effectively approach the rank of a matrix\cite{xie2016weighted,zha2020benchmark}.
	\label{definition1}
\end{definition}

\begin{remark}
    \textit{\textbf{Advantage of Utilizing the Schatten \( p \)-Norm Constraint}: 
    Consider $\bm{\mathcal{G}}$ as an example, where $\Delta^{(c)}$ denotes the $c$-th frontal slice of $\bm{\mathcal{G}}$. Let $\sigma_1, \sigma_2, \ldots, \sigma_h$ be the singular values of $\Delta^{(c)}$, ordered in descending sequence. The Schatten \( p \)-norm of $\Delta^{(c)}$ is defined as ${\left\| {\Delta^{(c)}} \right\|_{{\text{Sp}}}^p} = \sigma_1^p + \ldots + \sigma_h^p$. As $p \rightarrow 0$, the Schatten \( p \)-norm approximates the rank of $\Delta^{(c)}$. This is advantageous over the nuclear norm since minimizing the Schatten \( p \)-norm can more effectively ensure that $\Delta^{(c)}$ approximates a target rank, thus maintaining a low-rank spatial structure. Such a constraint facilitates the exploration of complementary information among different views, enhancing the convergence of labels across these views.}
\end{remark}

\begin{remark}
    \textit{Utilizing anchor points as intermediaries allows us to derive the transition probabilities from samples to categories. This is achieved by multiplying the transition probabilities from samples to anchor points with those from anchor points to categories. Consequently, we can establish a standardized soft label matrix $\mathbf{G}^{(v)}$, which acts as the cluster indicator matrix for the samples. These matrices serve as cluster indicators for both samples and anchor points. Despite potential differences in data distribution across views, the intrinsic geometric structure should remain consistent. In other words, the relationships between anchor points and categories in each view should be uniform. Thus, applying the Schatten \( p \)-norm constraint to the tensor $\mathcal{H}$ is beneficial in ensuring its low-rank structure.}
\end{remark}

\subsection{Optimization}
We employ the Augmented Lagrange Multiplier (ALM) method to address this problem. We introduce auxiliary variables $\mathbf{F}^{(v)}$, $\mathbf{Q}^{(v)}$, $\bm{\mathcal{J}}$, and $\bm{\mathcal{A}}$. By setting
$\mathbf{F}^{(v)} \geq 0$ and $\mathbf{Q}^{(v)}\geq 0,\;\mathbf{Q}^{(v)}\cdot \textbf{1} =\textbf{1}\;$, the model is reformulated as:

\begin{equation}
	\begin{aligned}	
		&\mathop{\min}\sum_{v=1}^{V}
		\frac{1}{\alpha^{(v)}} \|
		\mathbf{S}^{(v)}\mathbf{H}^{(v)} -
		\mathbf{G}^{(v)}\|_{F}^{2}+ \lambda_1\left\|\bm{\mathcal{J}}\right\|_{\Sp}^{p}+ \lambda_2\left\|\bm{\mathcal{A}}\right\|_{\Sp}^{p}
		\\ &+\frac{\mu_1}{2}\|
		\mathbf{G}^{(v)}-\mathbf{F}^{(v)}+\frac{\mathbf{Y_1}^{(v)}}{\mu_1}\|_{F}^{2}+\frac{\mu_2}{2}\|
		\mathbf{H}^{(v)}-\mathbf{Q}^{(v)}+\frac{\mathbf{Y_2}^{(v)}}{\mu_2}\|_{F}^{2}\\ 
		&+ \frac{\mu_3}{2}\|
		\bm{\mathcal{G}}-\bm{\mathcal{J}}+\frac{\bm{\mathcal{Y}_3}}{\mu_3}\|_{F}^{2}
		+ \frac{\mu_4}{2}\|
		\bm{\mathcal{H}}-\bm{\mathcal{A}}+\frac{\bm{\mathcal{Y}_4}}{\mu_3}\|_{F}^{2}\\ 
		&\text{s.t.}\; \mathbf{F}^{(v)}\geq 0,\;
		\mathbf{G}^{(v)^{\mathrm{T}}}\mathbf{G}^{(v)}=\mathbf{I},\;
		\mathbf{Q}^{(v)}\geq 0,\;\mathbf{Q}^{(v)}\cdot \textbf{1} =\textbf{1}\;\\
		&\qquad\alpha^{(v)} \geq 0,\; \sum_{v=1}^{V}\alpha^{(v)} = 1
	\end{aligned}	
	\label{optimization}
\end{equation}
where $\mathbf{Y_1}^{(v)},\mathbf{Y_2}^{(v)}, \bm{\mathcal{Y}}_3$, and $\bm{\mathcal{Y}}_4$ are the Lagrange multipliers, and $\mu_1, \mu_2, \mu_3, \mu_4$ serve as penalty parameters. The problem is decomposed into the following subproblems:

$\bullet$ With fixed values for
$\mathbf{H}^{(v)},\mathbf{Q}^{(v)},\mathbf{F}^{(v)},\bm{\mathcal{J}},\bm{\mathcal{A}},\text{and } \alpha^{(v)}$,
we aim to solve for $\mathbf{G}^{(v)}$. Equation (\ref{optimization}) can be expressed as:

\begin{equation}
	\begin{aligned}
		&\mathop{\min_{\mathbf{G}^{(v)}}}\sum_{v=1}^{V}
		\frac{1}{\alpha^{(v)}} \|
		\mathbf{S}^{(v)}\mathbf{H}^{(v)} -
		\mathbf{G}^{(v)}\|_{F}^{2} \\&+\frac{\mu_1}{2}\|
		\mathbf{G}^{(v)}-\mathbf{F}^{(v)}+\frac{\mathbf{Y_1}^{(v)}}{\mu_1}\|_{F}^{2} + \frac{\mu_3}{2}\|
		\bm{\mathcal{G}}-\bm{\mathcal{J}}+\frac{\bm{\mathcal{Y}_3}}{\mu_3}\|_{F}^{2}\\
		&\text{s.t.}\; \mathbf{G}^{(v)^{\mathrm{T}}}\mathbf{G}^{(v)}=\mathbf{I},
	\end{aligned}	
\end{equation}

By simplifying, we obtain the equivalent form of the equation:

\begin{equation}
	\begin{aligned}
		&\mathop{\max_{\mathbf{G}^{(v)}}}\sum_{v=1}^{V}tr\left( \mathbf{G}^{(v)^{\mathrm{T}}}\mathbf{B}^{(v)}\right) \\&\text{s.t.}\; \mathbf{G}^{(v)^{\mathrm{T}}}\mathbf{G}^{(v)}=\mathbf{I}
	\end{aligned}	
\end{equation}

Here,
\[\mathbf{B}^{(v)} = \frac{2}{\alpha^{(v)}}\mathbf{S^{(v)}}\mathbf{H}^{(v)}+\mu_1\left( \mathbf{F}^{(v)}-\frac{\mathbf{Y}_1^{(v)}}{\mu_1}\right)  + \mu_3\left( \mathbf{J}^{(v)}-\frac{\mathbf{Y}_3^{(v)}}{\mu_3}\right) .\]

To address this, we introduce the following theorem:

\begin{theorem}
	\label{orth}
	\cite{GaoXCDGL19} Given the compact singular value decomposition (SVD) of $\mathbf{W}$ as $\mathbf{U}\mathbf{\Sigma}\mathbf{V}^{\mathrm{T}}$, the optimal solution for
	\begin{equation}
		\mathop{\max_{\mathbf{M}^{\mathrm{T}}\mathbf{M} = \mathbf{I}}}tr\left( \mathbf{M}^{\mathrm{T}}\mathbf{W}\right) 
	\end{equation}
	is given by $\mathbf{M} = \mathbf{U}\mathbf{V}^{\mathrm{T}}$.
\end{theorem}

Based on Theorem \ref{orth}, the solution for $\mathbf{G}^{(v)}$ is:

\begin{equation}
	\label{update_G}
	\mathbf{G}^{(v)^*} = \mathbf{U}_{1}^{(v)}\mathbf{V}_{1}^{(v)^{\mathrm{T}}}
\end{equation}
with $\mathbf{U}_{1}^{(v)}$ and $\mathbf{V}_{1}^{(v)}$ derived from the singular value decomposition (SVD) of $\mathbf{B}^{(v)}$.

$\bullet$ With fixed terms
$\mathbf{G}^{(v)},\mathbf{Q}^{(v)}, \mathbf{F}^{(v)}, \bm{\mathcal{J}}, \bm{\mathcal{A}}, \alpha^{(v)}$,
the expression for $\mathbf{H}^{(v)}$ from (\ref{optimization}) is given by:

\begin{equation}
	\begin{aligned}
		&\mathop{\min_{\mathbf{H}^{(v)}}}\sum_{v=1}^{V}
		\frac{1}{\alpha^{(v)}} \|
		\mathbf{S}^{(v)}\mathbf{H}^{(v)} -
		\mathbf{G}^{(v)}\|_{F}^{2} \\
		&+\frac{\mu_2}{2}\|
		\mathbf{H}^{(v)}-\mathbf{Q}^{(v)}+\frac{\mathbf{Y_2}^{(v)}}{\mu_2}\|_{F}^{2} + \frac{\mu_4}{2}\|
		\bm{\mathcal{H}}-\bm{\mathcal{A}}+\frac{\bm{\mathcal{Y}_4}}{\mu_4}\|_{F}^{2}\\ 
		=& \mathop{\min_{\mathbf{H}^{(v)}}}\sum_{v=1}^{V} {tr(\mathbf{H}^{(v)\mathrm{T}}\mathbf{C}^{(v)}\mathbf{H}^{(v)})}-2tr(\mathbf{H}^{(v)\mathrm{T}}\mathbf{D}^{(v)})\\
	\end{aligned}
	\label{optimization_H}
\end{equation}
where$\mathbf{C}^{(v)} = \frac{1}{\alpha^{(v)}}\mathbf{S}^{(v)^{\mathrm{T}}}\mathbf{S}^{(v)}+(\frac{\mu_2}{2}+\frac{\mu_4}{2})I_m, \mathbf{D}^{(v)} =\frac{1}{\alpha^{(v)}}\mathbf{S}^{(v)^{\mathrm{T}}}\mathbf{G}^{(v)}$ $+\frac{\mu_2}{2}(\mathbf{Q}^{(v)} - \frac{\mathbf{Y}_2^{(v)}}{\mu_2} )+\frac{\mu_4}{2}(\mathbf{A}^{(v)} - \frac{\mathbf{Y}_4^{(v)}}{\mu_4} ) $.
Under unconstrained conditions, taking the derivative of the objective function and setting it to zero yields the solution for $\mathbf{H}^{(v)}$, i.e., $\mathbf{H}^{(v)^*}$ can be obtained by calculating as follows:
\begin{equation}
	\begin{aligned}
		\mathbf{C}^{(v)}\mathbf{H}^{(v)}- \mathbf{D}^{(v)} = 0
	\end{aligned}
	\label{optimization_H_1}
\end{equation}

Therefore, the solution for $\mathbf{H}^{(v)}$ is:
\begin{equation}
	\begin{aligned}
		\mathbf{H}^{(v)^*}= \mathbf{C}^{(v)^{-1}}\mathbf{D}^{(v)}
	\end{aligned}
	\label{update_H}
\end{equation}

$\bullet$ Given fixed values of $\mathbf{G}^{(v)},\mathbf{H}^{(v)},\mathbf{Q}^{(v)},\bm{\mathcal{J}},\bm{\mathcal{A}},$ and $\alpha^{(v)}$, we aim to solve for $\mathbf{F}^{(v)}$. Equation (\ref{optimization}) can be reformulated as:
\begin{equation}
	\begin{aligned}	
		\min_{\mathbf{F}^{(v)}}\frac{\mu_1}{2}\|\mathbf{G}^{(v)}-\mathbf{F}^{(v)}+\frac{\mathbf{Y}_1^{(v)}}{\mu_1}\|_{F}^{2}
		&= \min_{\mathbf{F}^{(v)}}\frac{\mu_1}{2}\|\overline{\mathbf{G}}^{(v)}-\mathbf{F}^{(v)}\|_{F}^{2}, \\
		&\text{s.t. } \mathbf{F}^{(v)} \geq 0,
	\end{aligned}	
	\label{F}
\end{equation}
where
\[
\overline{\mathbf{G}}^{(v)} = \mathbf{G}^{(v)} + \frac{\mathbf{Y}_1^{(v)}}{\mu_1}
\]
The solution to Equation (\ref{F}) is given by:
\begin{equation}
	\mathbf{F}^{(v)^*} = \max(\overline{\mathbf{G}}^{(v)}, 0)
	\label{update_F}
\end{equation}

$\bullet$ Given fixed values of $\mathbf{G}^{(v)},\mathbf{H}^{(v)},\mathbf{F}^{(v)},\bm{\mathcal{J}},\bm{\mathcal{A}},$ and $\alpha^{(v)}$, we aim to solve for $\mathbf{Q}^{(v)}$. Equation (\ref{optimization}) can be reformulated as:
\begin{equation}
	\begin{aligned}	
		&\min_{\mathbf{Q}^{(v)}}\frac{\mu_2}{2}\|\mathbf{H}^{(v)}-\mathbf{Q}^{(v)}+\frac{\mathbf{Y}_2^{(v)}}{\mu_2}\|_{F}^{2}=\min_{\mathbf{Q}^{(v)}}\frac{\mu_2}{2}\|\mathbf{Q}^{(v)}-\frac{\mathbf{K}^{(v)}}{\mu_2}\|_{F}^{2}\\
		&\text{s.t. } \mathbf{Q}^{(v)}\geq 0,\;\mathbf{Q}^{(v)}\cdot \textbf{1} =\textbf{1}\;\\
	\end{aligned}	
	\label{Q}
\end{equation}
where $\mathbf{K}^{(v)}=\mu_2\mathbf{H}^{(v)}+\mathbf{Y}_2^{(v)}$.
Following \cite{nie2016constrained}, the closed-form solution of $\mathbf{Q}^{(v)^*}$ is:
\begin{equation}
	\begin{aligned}	
		q_i^{(v^*)} = (\frac{\mathbf{K}^{(v)}_{i}}{\mu_2}+\gamma\textbf{1})_+
	\end{aligned}	
	\label{update_Q}
\end{equation}
where $\gamma$ is the Lagrangian multiplier.

$\bullet$ With the variables
$\mathbf{G}^{(v)},\mathbf{H}^{(v)},\mathbf{F}^{(v)},\mathbf{Q}^{(v)},\bm{\mathcal{A}},\alpha^{(v)}$ fixed,
to solve for $\bm{\mathcal{J}}$, equation (\ref{optimization}) is reformulated as:
\begin{equation}
	\mathop{\min_{\bm{\mathcal{J}}}} \lambda_1 \left\|\bm{\mathcal{J}}\right\|_{\Sp}^{p} + \frac{\mu_3}{2}\left\|
	\bm{\mathcal{G}}-\bm{\mathcal{J}}+\frac{\bm{\mathcal{Y}}_3}{\mu_3}\right\|_{F}^{2}
\end{equation}
Consequently, $\bm{\mathcal{J}}$ can be derived from:
\begin{equation}
	\bm{\mathcal{J}}_{*} = \arg \min  \frac{1}{2}\left\|
	\bm{\mathcal{J}}-(\bm{\mathcal{G}}+\frac{\bm{\mathcal{Y}}_3}{\mu_3})\right\|_{F}^{2}+\frac{\lambda_1}{\mu_3}\left\|\bm{\mathcal{J}}\right\|_{\Sp}^{p} 
\end{equation}
To address this, we present the subsequent theorem:
\begin{theorem}
	\label{tensor_theorem}
	\cite{gao2020enhanced} Given $\bm{\mathcal{Z}} \in \mathbb{R}^{n_1\times n_2\times n_3}$, the t-SVD of $\bm{\mathcal{Z}}$ is denoted as $\bm{\mathcal{Z}} = \bm{\mathcal{U}}*\bm{\mathcal{S}}*\bm{\mathcal{V}}^{\mathrm{T}}$. For the equation:
	\begin{equation}
		\mathop{\min_{\bm{\mathcal{X}}}} \frac{1}{2}\| \bm{\mathcal{X}}-\bm{\mathcal{Z}}\|_{F}^{2}+\tau\|\bm{\mathcal{X}}\|_{\Sp}^{p}
	\end{equation}
	the optimal solution is expressed as:
	\begin{equation}
		\bm{\mathcal{X}}^{*} = \Gamma_{\tau*n_3}(\bm{\mathcal{Z}}) = \bm{\mathcal{U}}*\text{ifft}(\mathbf{\textit{P}}_{\tau*n_3}(\overline{\bm{\mathcal{Z}}}))*\bm{\mathcal{V}}^{\mathrm{T}}
	\end{equation}
	where $\mathbf{\textit{P}}_{\tau*n_3}(\overline{\bm{\mathcal{Z}}})$ is a third-order tensor, and $\mathbf{\textit{P}}_{\tau*n_3}(\overline{\bm{\mathcal{Z}}}^{(i)})$ denotes the $\textit{i}_{th}$ frontal slice of this tensor. This can be computed using the \textit{Generalized soft-thresholding (GST)} method. 
\end{theorem}

In light of Theorem \ref{tensor_theorem}, the solution for $\bm{\mathcal{J}}$ is:
\begin{equation}
	\label{update_J}
	\bm{\mathcal{J}}^{*} = \Gamma_{\frac{\lambda_1}{\mu_3}*n_3}(\bm{\mathcal{G}}+\frac{\bm{\mathcal{Y}}_3}{\mu_3})
\end{equation}

$\bullet$ Given the fixed parameters:
$\mathbf{G}^{(v)}$, $\mathbf{H}^{(v)}$, $\mathbf{F}^{(v)}$, $\bm{\mathcal{J}}$, and $\alpha^{(v)}$,
we aim to solve for $\bm{\mathcal{A}}$. Equation (\ref{optimization}) can be reformulated as:

\begin{equation}
	\begin{aligned}	
		\mathop{\min_{\bm{\mathcal{A}}}} \left( \lambda_2 \|\bm{\mathcal{A}}\|_{\Sp}^{p} + \frac{\mu_4}{2} \left\| \bm{\mathcal{H}}-\bm{\mathcal{A}}+\frac{\bm{\mathcal{Y}}_4}{\mu_4} \right\|_{F}^{2} \right) \\=
		\mathop{\min_{\bm{\mathcal{A}}}} \left( \frac{1}{2} \left\| \bm{\mathcal{A}}-(\bm{\mathcal{H}}+\frac{\bm{\mathcal{Y}}_4}{\mu_4}) \right\|_{F}^{2} + \frac{\lambda_2}{\mu_4} \|\bm{\mathcal{A}}\|_{\Sp}^{p} \right)
	\end{aligned}	
\end{equation}

Similarly, the optimal solution for $\bm{\mathcal{A}}$ can be expressed as:

\begin{equation}
	\label{update_A}
	\bm{\mathcal{A}}^{*} = \Gamma_{\frac{\lambda_2}{\mu_4} \cdot n_3} \left( \bm{\mathcal{H}}+\frac{\bm{\mathcal{Y}}_4}{\mu_4} \right)
\end{equation}

$\bullet$ Given fixed matrices
$\mathbf{G}^{(v)},\mathbf{H}^{(v)},\mathbf{F}^{(v)},\bm{\mathcal{J}},\bm{\mathcal{A}}$,
we aim to solve for $\alpha^{(v)}$. In this context, the objective function is expressed as
\begin{equation}
	\begin{aligned}
		&\mathop{\min_{\alpha^{(v)}}}\sum_{v=1}^{V}
		\frac{1}{\alpha^{(v)}} \|\mathbf{S}^{(v)}\mathbf{H}^{(v)}-\mathbf{G}^{(v)}\|_{F}^{2}
		\\&\text{s.t.}\;\alpha^{(v)} \geq 0,\; \sum_{v=1}^{V}\alpha^{(v)} = 1
	\end{aligned}
\end{equation}

Let $\mathbf{T}^{(v)}=\|\mathbf{S}^{(v)}\mathbf{H}^{(v)}-\mathbf{G}^{(v)}\|_{F}^{2}$. Utilizing the Lagrange multiplier method, the optimal $\alpha^{(v)}$ can be computed as:
\begin{equation}
	\alpha^{(v)} = \frac{\sqrt{\mathbf{T}^{(v)}}}{\sum_{v=1}^{V}\sqrt{\mathbf{T}^{(v)}}}
	\label{update_alpha}
\end{equation}

Furthermore, the Lagrange multipliers $\mathbf{Y}_1^{(v)},\mathbf{Y}_2^{(v)}, \bm{\mathcal{Y}}_{3}, \bm{\mathcal{Y}}_4$ are updated by:
\begin{equation}
	\label{Lm}
	\begin{aligned}
		&\mathbf{Y}_1^{(v)} = \mathbf{Y}_1^{(v)} + \mu_1(\mathbf{G}^{(v)}-\mathbf{F}^{(v)})\\
		&\mathbf{Y}_2^{(v)} = \mathbf{Y}_2^{(v)} + \mu_2(\mathbf{H}^{(v)}-\mathbf{Q}^{(v)})\\
		&\bm{\mathcal{Y}}_{3} = \bm{\mathcal{Y}}_{3} + \mu_3(\bm{\mathcal{G}}-\bm{\mathcal{J}})\\
		&\bm{\mathcal{Y}}_{4} = \bm{\mathcal{Y}}_{4} + \mu_4(\bm{\mathcal{H}}-\bm{\mathcal{A}})
	\end{aligned}
\end{equation}

The penalty parameters $\mu_i, i=1,2,3,4$, are updated as:
\begin{equation}
	\label{pp}
	\begin{aligned}
		\mu_i = \min(\mu_i \times 1.1, max\_\mu)
	\end{aligned}
\end{equation}
where $max\_\mu$ are predefined constants.

To conclude, we compute a shared matrix $\mathbf{G}$ using
\begin{equation}
	\mathbf{G} = \frac{\sum_{v=1}^{V} \frac{\mathbf{G}^{(v)}}{\alpha^{(v)}}}{\sum_{v=1}^{V} \frac{1}{\alpha^{(v)}}}
\end{equation}
for all views. The position of the maximum value in each row corresponds to the label for that sample.

The complete algorithm is detailed in Algorithm \ref{alg1}.

\begin{algorithm}
	\caption{Algorithm for OSMVC-TP}
	\label{alg1}
	\begin{algorithmic}[1] %[1] enables line numbers
		\STATE \textbf{Input}: Original data set $\{\textbf{X}^{(v)} \}_{v=1}^{V} \in \mathbb{R}^{N \times d}$.\\
		\STATE \textbf{Parameter}: Anchor rate, parameters $\lambda_1$, $\lambda_2$, $p$.\\
		\STATE \textbf{Output}: Clustering labels.\\
		\STATE Construct anchor graphs $\mathbf{S}^{(v)} \in \mathbb{R}^{n \times m}$ using.
		\STATE Initialize $\mathbf{F}^{(v)}=\mathbf{Q}^{(v)}=\mathbf{Y}_1^{(v)}=\mathbf{Y}_2^{(v)}=0$, $\bm{\mathcal{J}}=\bm{\mathcal{Y}_3}=\bm{\mathcal{A}}=\bm{\mathcal{Y}_4}=0$, $\mu_i =10^{-3}$, $ max\_\mu = 10^9 $, $\eta = 1.1$, $\alpha^{(v)} = \frac{1}{V}$.
		\WHILE{not converged}
		\STATE Update $\mathbf{G}^{(v)}$ using (\ref{update_G}).
		\STATE Update $\mathbf{H}^{(v)}$ using (\ref{update_H}).
		\STATE Update $\mathbf{F}^{(v)}$ using (\ref{update_F}).
		\STATE Update $\mathbf{Q}^{(v)}$ using (\ref{update_Q}).
		\STATE Update $\bm{\mathcal{J}}$ using (\ref{update_J}).
		\STATE Update $\bm{\mathcal{A}}$ using (\ref{update_A}).
		\STATE Update $\alpha^{(v)}$ using (\ref{update_alpha}).
		\STATE Update $\mathbf{Y}_1^{(v)}, \mathbf{Y}_2^{(v)}, \bm{\mathcal{Y}}_{3}, \bm{\mathcal{Y}}_4$ using (\ref{Lm}).
		\STATE Update penalty parameters $\mu_i, i=1,2,3,4$ using (\ref{pp}).
		\STATE Compute indicator matrix $\mathbf{G} = \left(\sum\frac{\mathbf{G}^{(v)}}{\alpha^{(v)}}\right) / \left(\sum \frac{1}{\alpha^{(v)}}\right)$ and extract the corresponding labels.
		\ENDWHILE
		\STATE \textbf{return} Clustering labels.
	\end{algorithmic}	
\end{algorithm}
\section{Experiment}
In this section, we evaluate the efficacy of our proposed method through comprehensive experiments. All experiments are conducted on a standard Windows 10 Server, equipped with dual Intel(R) Xeon(R) Gold 6230 CPUs and 128 GB RAM.

\subsection{Datasets and Metrics}
% Description of the experimental setup
Our experiments were conducted on six widely-recognized datasets. This approach was chosen to rigorously validate the effectiveness of our model. Detailed information about each dataset is provided in Table \ref{table_datasets}, where we present key characteristics and statistics.
\begin{table}[!t]
	\caption{Statistics of Real Benchmark Datasets}
	\label{table_datasets}
	\begin{center}
		\resizebox{\columnwidth}{!}{
			\begin{tabular}{c!{\vrule width 1pt}cccc!{\vrule width 1pt}cc}
				\toprule
				Scale &\multicolumn{4}{c!{\vrule width 1pt}}{Normal}&\multicolumn{2}{c}{Large}\\
				\midrule
				Dataset & MSRC & HW4 & Mnist &Scene15& Reuters& NoisyMNIST\\
				\midrule
				Size& 210 & 2000 & 4000 & 4485 & 18758 & 50000\\
				Views& 5 & 4 & 3 & 3&  5  & 2\\
				Clusters& 7 & 10  & 4 & 15 & 6 & 10\\
				\midrule
		\end{tabular}}
	\end{center}
\end{table}

We employ the following metrics to gauge clustering performance:
(1) ACC; (2) NMI; (3) Purity.
Higher scores in these indices signify superior model performance.

\subsection{Compared methods}
\begin{itemize}
  \item \textbf{CSMSC} \cite{luo2018consistent}: This method divides the self-representation coefficient matrix for each view into two components: consistencies, which demonstrate a low-rank structure, and specificities, which highlight the unique variations in each view.
  \item \textbf{GMC} \cite{wang2019gmc}: Integrates graph learning that is both view-consistent and view-specific within a unified framework, allowing for joint optimization.
  \item \textbf{ETLMSC} \cite{wu2019essential}: Constructs a probability matrix from a tensor and employs spectral clustering to produce the final results.
  \item \textbf{LMVSC} \cite{kang2020large}: Employs a multi-graph fusion strategy to create a consistent bipartite graph, followed by spectral clustering to determine cluster labels.
  \item \textbf{FMCNOF} \cite{yang2020fast}: Offers a fast clustering approach using NMF (Non-negative Matrix Factorization) and anchor selection.
  \item \textbf{SFMC} \cite{9146384}: Combines Laplace rank constraints with a bipartite graph learning strategy to create view-consistent bipartite graphs.
  \item \textbf{MSC-BG} \cite{yang2022multiview}: Utilizes the \textbf{\textit{Schatten p}}-norm to constrain the bipartite graph, effectively capturing the spatial structure and complementary information in the views.
  \item \textbf{FPMVS-CAG} \cite{wang2021fast}: Unifies anchor learning and graph construction, ensuring joint optimization with linear time complexity.
\end{itemize}

\begin{table*}[!t]
	\centering
	\caption{The results on the MSRC, HW4, Mnist4 and Scene15 datasets.}
	\label{result1}
	\setlength\tabcolsep{6pt}
	\singlespacing
	\begin{tabular}{ c| c c c |c c c |c c c |c c c }
		\toprule
		Datasets & \multicolumn{3}{c|}{MSRC} &\multicolumn{3}{c|}{HW4}& \multicolumn{3}{c|}{Mnist4} & \multicolumn{3}{c}{Scene15} \\
		Methods & ACC & NMI & PUR & ACC & NMI & PUR & ACC & NMI & PUR & ACC & NMI & PUR  \\
		\midrule
		CSMSC & 0.862 & 0.767 & 0.862 & 0.806 & 0.793 & 0.867 & 0.643 & 0.601 & 0.728 & 0.576 & 0.574 & 0.629  \\
		GMC & 0.895 & 0.809 & 0.895 & 0.879 & 0.882 & 0.879 & 0.921 & 0.807 & 0.921 & 0.409 & 0.43 & 0.417  \\
		ETLMSC & 0.962 & 0.937 & 0.962 & 0.938 & 0.894 & 0.938 & 0.934 & 0.847 & 0.934 & 0.218 & 0.166 & 0.221  \\
		LMVSC & 0.814 & 0.717 & 0.814 & 0.904 & 0.831 & 0.904 & 0.892 & 0.726 & 0.892 & 0.561 & 0.512 & 0.581  \\
		FMCNOF & 0.713 & 0.648 & 0.714 & 0.541 & 0.484 & 0.54 & 0.686 & 0.513 & 0.695 & 0.263 & 0.249 & 0.258  \\
		SFMC & 0.809 & 0.721 & 0.781 & 0.853 & 0.871 & 0.873 & 0.917& 0.801 & 0.917 & 0.188&0.135 &0.202\\
		MSC-BG & 0.981 & 0.960 & 0.981 & 0.889 & 0.922 & 0.889 & 0.938 & 0.861 & 0.938 & 0.519 & 0.602 & 0.562\\
		FPMVS-CAG & 0.843 & 0.738 & 0.843 & 0.850 & 0.787 & 0.850 & 0.887 & 0.719 & 0.887 & 0.541 & 0.584 & 0.541  \\
		%		\textbf{Ours}Q-non-orth
		\textbf{Ours} & \textbf{1.000} & \textbf{1.000} & \textbf{1.000}& \textbf{0.996}  &  \textbf{0.989} & \textbf{0.996} &  \textbf{0.991} & \textbf{0.965} & \textbf{0.991} & \textbf{0.853} & \textbf{0.892} & \textbf{0.893}\\
	    \bottomrule
	\end{tabular}
\end{table*}

\begin{table}[!t]
	\centering
	\caption{The results on the Reuters, and NoisyMNIST datasets. 'OM' means Out of Memory. "-" means takes more than 3 hours to calculate. }
	\label{result2}
	\setlength\tabcolsep{4.5pt}
	\singlespacing
	\begin{tabular}{ c | c c c | ccc}
		\toprule
		Datasets & \multicolumn{3}{c|}{Reuters}& \multicolumn{3}{c}{NoisyMNIST}  \\
		Methods & ACC & NMI & PUR & ACC & NMI & PUR \\
		\midrule
		CSMSC & OM & OM & OM  & OM & OM & OM\\
		GMC & - & - & -  & - & - & - \\
		ETLMSC & OM & OM & OM & OM & OM & OM \\
		LMVSC & 0.589 & 0.335 & 0.615 & 0.388 & 0.344 & 0.434\\
		FMCNOF & 0.343 & 0.125 & 0.358 & 0.333 & 0.237 & 0.340\\
		SFMC & 0.602 & 0.354 & 0.604 & 0.699 & 0.681 & 0.727\\
		MSC-BG & 0.640 & 0.484 & 0.686 & - & - &-\\
		FPMVS-CAG & 0.526 & 0.323 & 0.603 & 0.554 & 0.513 & 0.567\\
		\textbf{Ours} & \textbf{0.819} & \textbf{0.699} & \textbf{0.834} & \textbf{0.787} & \textbf{798}&\textbf{0.820} \\
		\bottomrule
	\end{tabular}
\end{table}
To guarantee the precision and impartiality of the experimental results, we report the mean values derived from ten iterations, each conducted with optimal parameters. The corresponding experimental data are systematically tabulated in Table \ref{result1} and Table \ref{result2}.

\subsection{Experimental results}
Tables \ref{result1} and \ref{result2} present three key metrics evaluating our method on four small-scale and two large-scale datasets. Detailed analysis of these tables shows that our approach significantly outperforms existing methods in all six datasets. Specifically, FMCNOF efficiently acquires the soft label matrix via NMF on the anchor graph. However, it mandates consistent labels across different views, ignoring the potential complementary information between them. In contrast, our method employs the \textit{Schatten p}-norm constraint to progressively align the soft label matrices of various views. This strategy effectively harnesses the complementary information, leading to enhanced clustering accuracy. 

Moreover, while MSC-BG integrates tensor constraints, it requires intricate parameter tuning during the learning phase of k distinct connected components. On the other hand, the CSMSC and ETLMSC methods, employing a two-stage clustering approach, may encounter memory issues with large datasets, rendering them less suitable for large-scale clustering tasks. In contrast, our method mitigates such challenges by implementing anchor selection, significantly reducing computational burdens. This approach allows our method to not only efficiently process but also yield impressive results on large-scale datasets, as evidenced by its performance on the Reuters and NoisyMNIST datasets.

\subsection{Ablation study}
\begin{table*}[!t]
	\centering
	\setlength\tabcolsep{6pt}
	\singlespacing
	\caption{The results of ablation study.}
	\label{ablation}
	\begin{tabular}{c|c c c|c  c c| c c c |c c c}
		\toprule
		Datasets & \multicolumn{3}{c|}{MSRC} & \multicolumn{3}{c|}{HW4} &\multicolumn{3}{c|}{Mnist4} & \multicolumn{3}{c}{Scene15}  \\
		Methods & ACC & NMI & PUR & ACC & NMI & PUR & ACC & NMI & PUR & ACC & NMI & PUR  \\
		\midrule
		w.o. $\bm{\mathcal{G}}\& \bm{\mathcal{H}}$ & 0.442 & 0.298 & 0.447 & 0.296 & 0.233 & 0.311 &0.368 & 0.123 & 0.424 & 0.092 & 0.004 & 0.069  \\
		w.o. $\bm{\mathcal{G}}$ & 0.914 & 0.829 & 0.914 & 0.661 & 0.573 & 0.682 & 0.916 & 0.790 & 0.916 & 0.387 & 0.334 & 0.397 \\
		w.o. $\bm{\mathcal{H}}$ & 0.990 & 0.978 & 0.990 & 0.980 & 0.958 & 0.980 & 0.978 & 0.926 & 0.978 & 0.755 & 0.800 & 0.791  \\
		\textbf{Ours} & \textbf{1.000} & \textbf{1.000} & \textbf{1.000}& \textbf{0.996}  &  \textbf{0.989} & \textbf{0.996} &  \textbf{0.991} & \textbf{0.965} & \textbf{0.991} & \textbf{0.853} & \textbf{0.892} & \textbf{0.893}\\
		\midrule
	\end{tabular}
\end{table*}

\begin{figure*}
	\centering
	\subfigure[Iteration 1 (ACC = 0.101)]{
		\centering
		\includegraphics[width=0.3\linewidth]{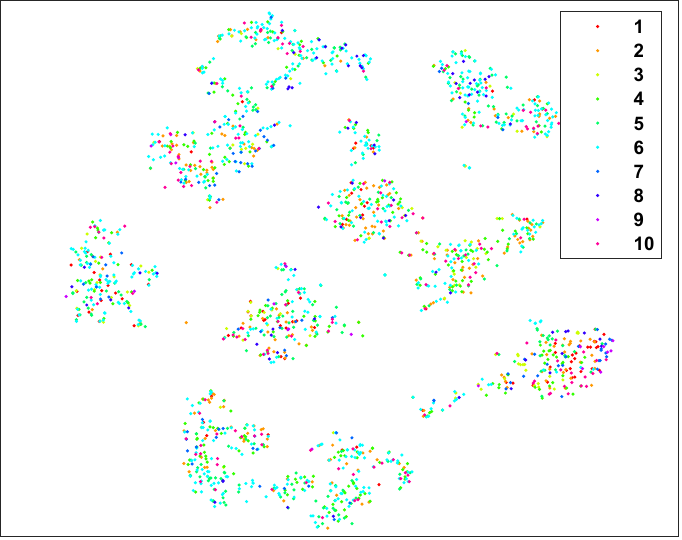}
	}
	\subfigure[Iteration 75 (ACC = 0.459)]{
		\centering
		\includegraphics[width=0.3\linewidth]{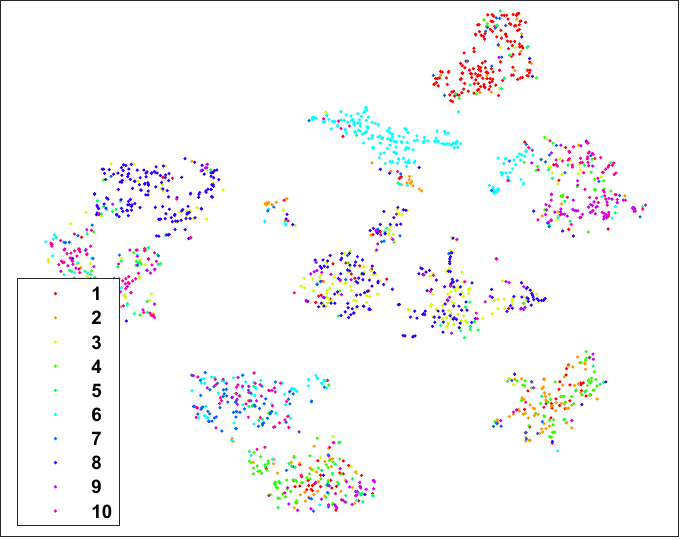}
	}
	\subfigure[Iteration 150 (ACC = 0.998)]{
		\centering
		\includegraphics[width=0.3\linewidth]{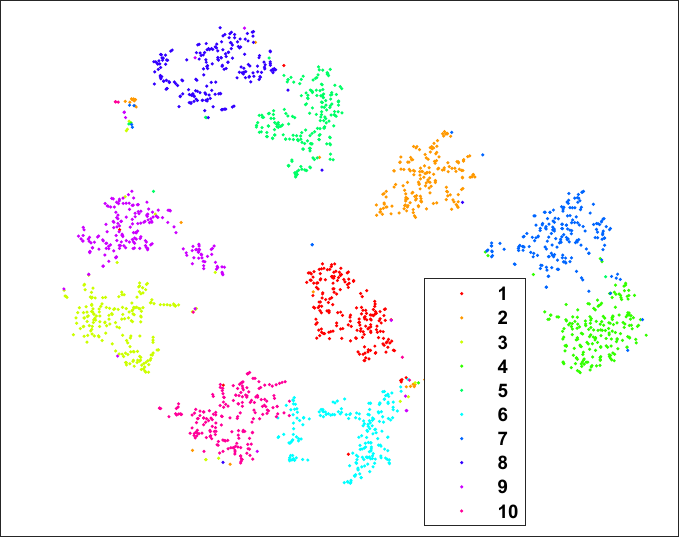}
	}
	\centering
	\caption{The clustering performances expressed by t-SNE.}
	\label{result_tsne}
\end{figure*}

To examine the significance of the two regularization terms, we performed ablation studies by omitting the two \textit{Schatten p}-norms on four datasets. The outcomes are presented in Table \ref{ablation}. Observations reveal that the performance is $50\%$ below when directly calculating the transition probability from samples to categories to learn labels than our proposed method. This may be because there are significant differences in data distribution among different views. Because the labels of samples and anchor points are consistent across different views, we can utilize tensor low-rank constraints to explore the complementary information among views. During the optimization process, this can gradually align the labels among different views and make them more consistent. Incorporating both \textit{Schatten p}-norms enhances the clustering outcome, with the term $\|\bm{\mathcal{G}}\|_{\Sp}$ having a more pronounced impact. Leveraging tensors helps capture complementary information between views, resulting in more accurate clustering labels.

\begin{figure}
	\centering
	{\subfigure[MSRC]{
			\centering
			\includegraphics[width=0.47\linewidth]{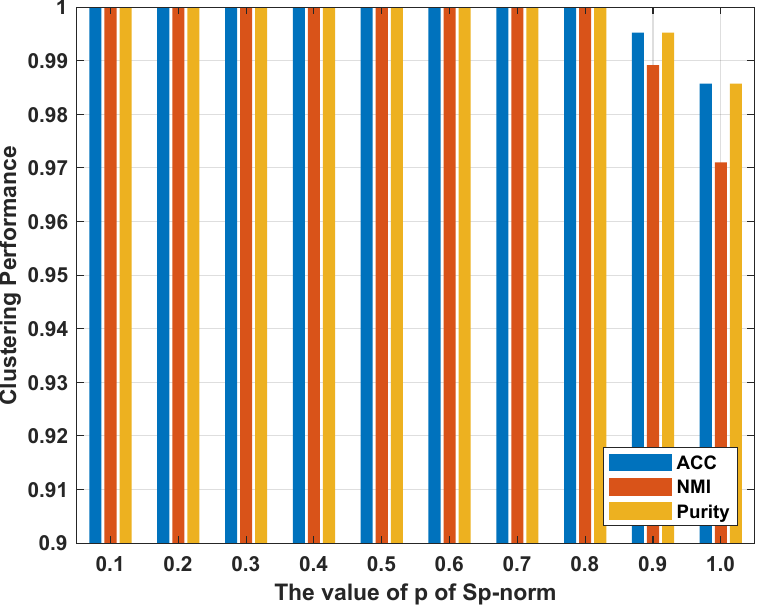}
		}
		\subfigure[HW4]{
			\centering
			\includegraphics[width=0.47\linewidth]{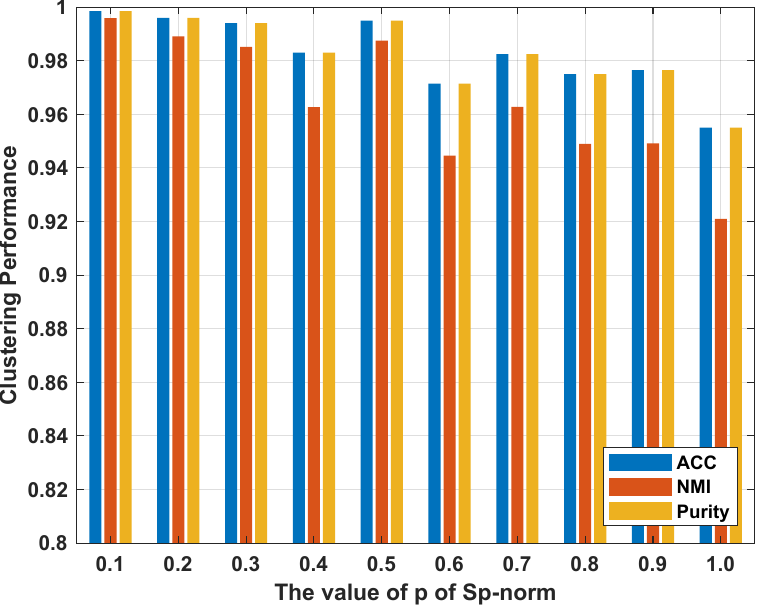}
	}}
	\centering
	\caption{Clustering performance vs. $p$ on four datasets.}
	\label{result_p}
\end{figure}

\subsection{Effect of parameter p}
We investigate the influence of the parameter $p$ on the \textbf{\textit{Schatten p}}-norm constraint across two different datasets. To ensure consistency, we maintain the same value of $p$ for both $\|\bm{\mathcal{G}}\|_{\Sp}^p$ and $\|\bm{\mathcal{H}}\|_{\Sp}^p$. We varied $p$ within the range of $0.1$ to $1.0$, and the corresponding results are shown in Fig. \ref{result_p}. Overall, the model achieves higher metrics when $p < 1$ compared to $p=1$. The \textbf{\textit{Schatten p}}-norm, with $p < 1$, enforces a spatially low-rank structure for the tensor's tangent plane, allowing for a more effective extraction of complementary information from multiple views. As a result, consistent cluster indicator matrices can be obtained from each view.

\begin{figure}
	\centering
	\subfigure[MSRC]{
		\centering
		\includegraphics[width=0.47\linewidth]{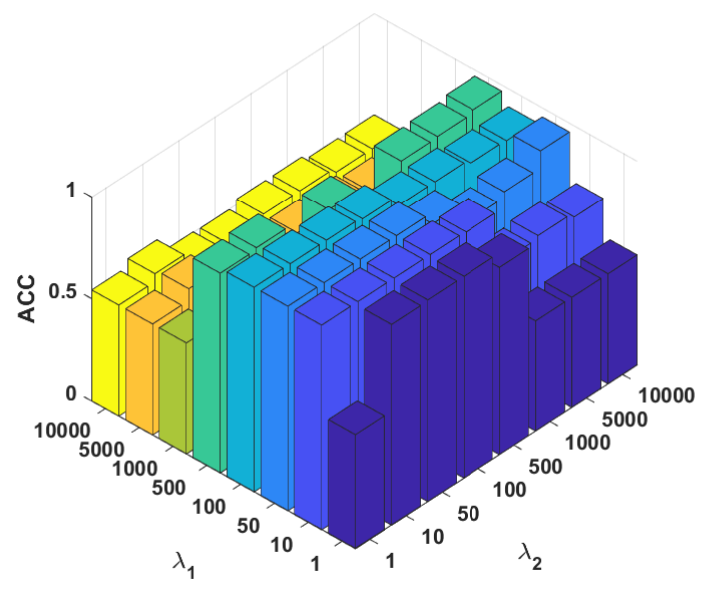}
	}
	\subfigure[HW4]{
		\centering
		\includegraphics[width=0.47\linewidth]{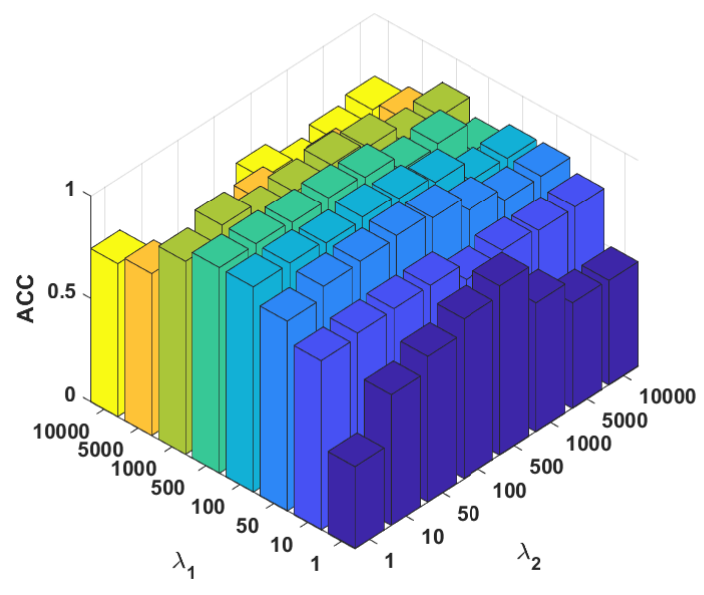}
	}
	\centering
	\caption{Clustering performance vs. $\lambda_1$ and $\lambda_2$ on two datasets.}
	\label{result_lambda}
\end{figure}

\subsection{Effect of lambda}
To examine the influence of the hyperparameters $\lambda_1$ and $\lambda_2$ on clustering performance, we conducted a fine-tuning process over the set $\{1, 10, 50, 100, 500, 1000, 5000, 10000\}$. The results for the two datasets are presented in Fig.~\ref{result_lambda}. The selection of these hyperparameters is crucial in determining the quality of the clustering results. Extreme values, whether too high or too low, will harm clustering performance. Our experiments reveal that fine-tuning both hyperparameters within the range of $[50, 500]$ leads to stable performance with minimal variations. The best performance is achieved when the two hyperparameters are configured to strike an optimal balance between the two components of the total loss.

\subsection{T-SNE results analysis}
T-SNE is a dimensionality reduction technique that preserves the relative distance relationships of the original data, allowing for visualization of data distribution characteristics. In this study, we present the t-SNE visualization results of the raw data based on the learned labels with varying numbers of iterations on the Handwritten4 dataset. The results are shown in Fig.~\ref{result_tsne}. Initially, most of the samples are grouped, indicating poor clustering performance. However, as the number of iterations increases, the boundaries between clusters become more distinct. Samples belonging to the same cluster start to gather together, leading to a gradual improvement in clustering accuracy. By the 150th iteration, the model has converged and successfully separated the data into 10 distinct clusters. The clustering accuracy reaches 0.998, providing strong evidence of the effectiveness of the model.

\begin{figure}
	\centering
	\subfigure[MSRC]{
		\centering
		\includegraphics[width=0.47\linewidth]{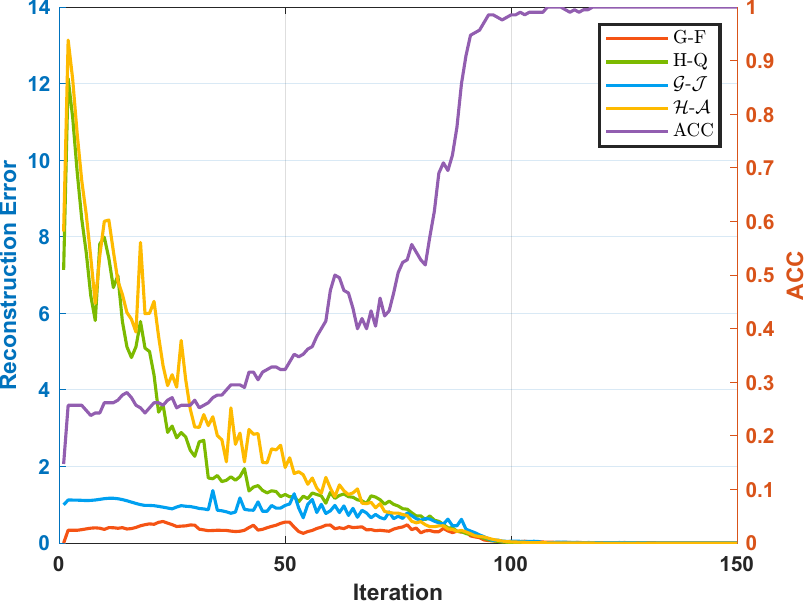}
	}
	\subfigure[HW4]{
		\centering
		\includegraphics[width=0.47\linewidth]{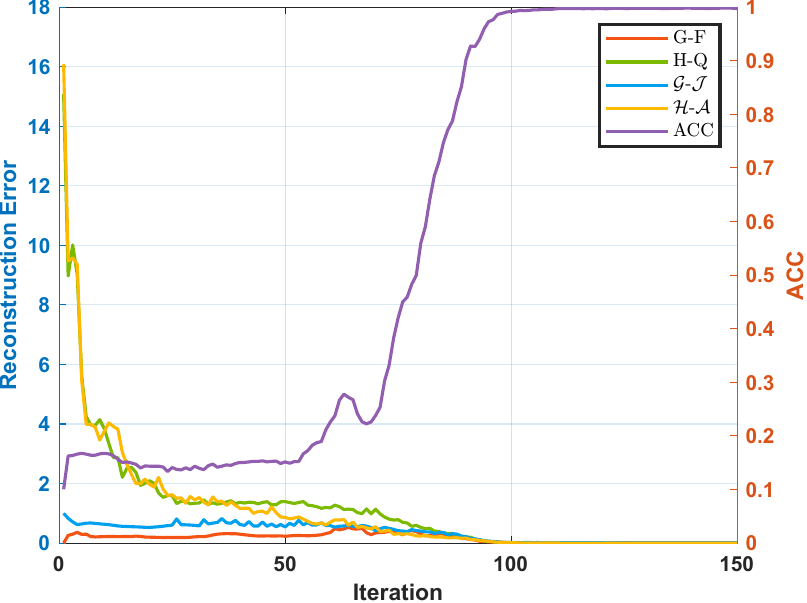}
	}
	\centering
	\caption{Convergence and clustering performance on two datasets.}
	\label{result_converge}
\end{figure}

\subsection{Convergence analysis}
We monitor the changes in reconstruction errors (i.e., $\|\mathbf{G}^{(v)}-\mathbf{F}^{(v)}\|_\infty$, $\|\mathbf{H}^{(v)}-\mathbf{Q}^{(v)}\|_\infty$, $\|\bm{\mathcal{G}} - \bm{\mathcal{J}}\|_\infty$, and $\|\bm{\mathcal{H}}-\bm{\mathcal{A}}\|_\infty$) over increasing iterations on two datasets. It is worth noting that the errors exhibit significant fluctuations during the initial 100 iterations but eventually stabilize. In our model, convergence is typically achieved after around 120 iterations.

The clustering accuracy is depicted as the number of iterations increases. During the initial 110 iterations, the accuracy experiences fluctuations at lower values as the model needs to converge. However, around the 105th iteration, the accuracy reaches its peak and remains relatively stable, with only slight fluctuations. Once the errors converge, the performance becomes stable.

\section{Conclusion}
This article offers a probabilistic interpretation of anchor graphs, portraying them as representations of the likelihood of sample transitions to anchor points. Capitalizing on the concept of transition probability, the article introduces a novel approach to deduce the transition probability matrix from anchors to categories. This matrix acts as a soft label matrix for anchor points. Concurrently, it directly computes the soft label matrix for the samples, basing on the transfer probabilities from samples to clusters. This methodology facilitates the acquisition of labels for samples in a single step. The application of the \textit{Schatten p}-norm significantly augments the extraction of complementary information across multiple views, thereby amplifying the clustering performance. A series of comprehensive experiments on datasets varying in scale substantiate the effectiveness of this approach.

\newpage
\bibliographystyle{ACM-Reference-Format}
\bibliography{sample-base}

\end{document}